\documentclass{article}

\usepackage{PRIMEarxiv}
\usepackage{natbib}
\usepackage[utf8]{inputenc} % allow utf-8 input
\usepackage[T1]{fontenc}    % use 8-bit T1 fonts
\usepackage{hyperref}       % hyperlinks
\usepackage{url}            % simple URL typesetting
\usepackage{booktabs}       % professional-quality tables
\usepackage{amsfonts}       % blackboard math symbols
\usepackage{nicefrac}       % compact symbols for 1/2, etc.
\usepackage{microtype}      % microtypography
\usepackage{lipsum}
\usepackage{fancyhdr}       % header
\usepackage{graphicx}       % graphics
\graphicspath{{media/}}     % organize your images and other figures under media/ folder

\usepackage{amsmath}
\usepackage{amsthm}      % For definition environment
\usepackage{xcolor}
\usepackage[table]{xcolor} 
\usepackage{colortbl}
\usepackage{bbm}
\usepackage{multirow}

\title{Discovering the Hidden Role of Gini Index In Prompt-based Classification}
\author{
  Ruixi Lin \\
  Independent Researcher\\
  \texttt{\{ruixi\}@u.nus.edu} \\
}

\begin{document}
\maketitle

\begin{abstract}
% 1. gini index启发了设么具体问题？比如：为什么现有方法在x场景失效？
% 2. gini 引导你发现了什么新现象/机制/设计原则？这是“discover”的核心！
% 3. 如果没有这个概念，你会忽略什么？

% 痛点：当前对 [XX 现象] 的理解受限于 [传统视角]，忽略了 [关键维度]。  
% 引入概念：我们提出将 [你的启发性概念] 作为新透镜，它强调 [核心思想]。  
% 你的发现：通过 [方法，哪怕只是 qualitative analysis / case study]，我们发现 [具体 insight]，例如 [举一个反直觉的例子]。  
% 价值：这不仅解释了 [长期困惑的现象]，还为 [设计/评估/理论] 提供了新原则。

In classification tasks, the long-tailed minority classes usually offer the predictions that are most important. Yet these classes consistently exhibit low accuracies, whereas a few high-performing classes dominate the game. We pursue a foundational understanding of the hidden role of \textit{Gini Index} as a tool for detecting and optimizing (debiasing) disparities in class accuracy, focusing on the case of prompt-based classification. We introduce the intuitions, benchmark Gini scores in real-world LLMs and vision models, and thoroughly discuss the insights of Gini not only as a measure of relative accuracy dominance but also as a direct optimization metric. Through rigorous case analyses, we first show that weak to strong relative accuracy imbalance exists in both prompt-based, text and image classification results and regardless of whether the classification is high-dimensional or low-dimensional. Then, we harness the Gini metric to propose a post-hoc model-agnostic bias mitigation method. Experimental results across few-shot news, biomedical, and zero-shot image classification show that our method significantly reduces both relative and absolute accuracy imbalances, minimizing top class relative dominance while elevating weakest classes.
\end{abstract}

% keywords can be removed

\section{Introduction}
% 1.1 分类任务：语言、图像等
% 1.2 Long-tail 导致的 Gini 高现象
% 1.3 不均衡原因：原始数据
% 1.4 重要性：小众样本-分类的重要程度
% 1.5 在训练数据层面解决的高成本
% 1.6 本质的输出向量不准确，修正分类输出向量的途径

% — A Study in Human-AI Collaboration” 或 “— Evidence from Large-Scale Code Repositories

% The Long-tail Issue In Prompt-Based Classification Outputs

% long-tail distribution in the completion string in Prompt-Based Classification Outputs\citep{todo}...

Classification tasks form the backbone of modern artificial intelligence, spanning diverse domains from natural language processing to computer vision. Whether classifying text into categories, recognizing objects in images, or assigning labels to multi-modal inputs, the fundamental challenge remains the same: models must learn to correctly distinguish between classes based on training data. In recent years, large language models (LLMs) and vision-language models have achieved remarkable performance across these tasks, yet they often struggle with a persistent problem---accuracy imbalance across classes---whose causes are rooted in the pretraining data itself.

A pervasive phenomenon in real-world data is the long-tailed distribution, where a small number of ``head'' classes occupy the majority of the dataset while the vast majority of ``tail'' classes have very few samples. In such cases, the performance of deep learning models is often dominated by the head classes while the learning of the tail classes is severely underdeveloped. This imbalance manifests directly in classification accuracy: models become biased toward frequently seen classes, while rare classes suffer from poor recognition performance. Though the exact classes may not be seen in pretraining, LLM completions can be degenerate, carrying more weight on frequently seen tokens \citep{chang2022}.

% Though the exact classes may not be seen in pretraining, LLM completions can be degenerate, carrying more weight on frequently seen tokens due to mechanisms such as induction heads that amplify repeated patterns \citep{chang-bergen-2022-tacl, induction-head-2025}.

Yet the importance of these minority classes cannot be overstated. In many critical applications---medical diagnosis, fraud detection, anomaly identification, and scientific discovery---the rare categories often carry the highest stakes. A model that performs well on common cases but fails on rare ones may be practically useless or even dangerous. The challenge, therefore, is not merely achieving high average accuracy, but ensuring adequate performance across all classes, particularly those with limited representation.

Traditional approaches to addressing class imbalance have focused primarily on the training data level. Methods incorporating oversampling, undersampling, and data augmentation aim to rebalance the dataset, followed by training with imbalance-aware loss functions \citep{cao2019,cui2019,buda2018}. However, iterative retraining or fine-tuning on rebalanced datasets becomes prohibitively expensive. Moreover, the cost of collecting and annotating sufficient high-quality tail-class examples to achieve natural balance is often high.

This suggests a fundamental shift in perspective: rather than fixing imbalance at the data level, we might address it at the output level. The core problem is not merely that training data is imbalanced, but that this imbalance produces systematically distorted output vectors---class predictions that favor head classes and suppress tail classes. If we can detect and correct this distortion in the model's predictions themselves, we may achieve better class fairness without the prohibitive cost of data rebalancing. \textbf{This points to the need for metrics that can quantify output-level imbalance and methods that can post-hoc optimize for more equitable predictions across classes}.

In this paper, we reveal one such metric---the Gini Index. We carefully examine Gini's hidden role as a tool for detecting and optimizing (debiasing) disparities in class accuracy, focusing on the case of prompt-based classification. In particular, we first empirically demonstrate Gini values in representative LLM-based text and image classification scenarios, and then we propose a post-hoc model-agnostic bias mitigation method to reduce Gini. We show that:
\begin{itemize}
    \item As measured by the Gini index, \textit{weak to strong relative accuracy imbalance} exists in both prompt-based, text and image classification results and regardless of whether the classification is high-dimensional or low-dimensional. Crucially, \textbf{the non-zero Gini (Gini > 0) demonstrates that \textit{relative accuracy imbalance} is present in real-world models}. \\
    \item Reducing relative accuracy imbalance is possible and meaningful. We achieve this by directly leveraging the Gini index as an optimization metric in \textbf{a post-hoc model-agnostic bias mitigation method}. \\
    \item \textbf{Both relative and absolute accuracy imbalance are mitigated} by our Gini-based bias mitigation method.
\end{itemize}

% ===OR
% While the term ‘[keyword]’ has recently surfaced in unpublished submissions (e.g., Author et al., 2025, OpenReview: [link]), its application remains ambiguous and context-sensitive—motivating our effort to clarify its conceptual role in [your domain].

\section{Related Work}
\label{sec:relatedwork}

\subsection{The Inequality Measure: Gini Index}
The Gini index, or Gini coefficient, is the most commonly used measure of inequality \citep{gini1912,hirschman1964}. It was defined based on the Lorenz curve \citep{lorenz,gastwirth1972}, ranging from 0 to 1, with 0 indicating perfect equality and 1 indicating perfect inequality.

The Gini index can be estimated using various approaches, including direct and indirect calculation methods \citep{heshmati2004}. In practice, when measuring income inequality, the choice between direct and indirect Gini index calculation methods is more than just convenience, but statistical robustness under different data conditions \citep{cowell2000}.

For example, the income Gini adopts a direct calculation method, which is directly computed from individual or grouped data without assuming an underlying functional form for the income distribution \citep{yao1999}. Income Gini for $N$ individuals (their income is denoted by $I$, average income is denoted by $\overline{\gamma}^{\text{I}}$) is defined as:
\begin{equation}
    G^{\text{Income}} = \frac{ \sum_{i=1}^{N} \sum_{j=1}^{N} |I_i - I_j| }{ 2N^2 \overline{\gamma}^{\text{I}} }
\end{equation}
Intuitively, income Gini highlights top-class relative dominance to the mean. As we will see later when formally defining the Gini index, this direct reciprocal calculation used in income Gini fits well in the context of measuring class accuracy disparity.

% The calculation of Gini index takes various forms \citep{todo}, including direct and indirect calculation methods. The appropriate method often depends on the nature of the available data---whether it is individual-level data or grouped into intervals.

\subsection{Prompt-based Classification}
LLMs can be viewed as modeling the probability distribution of completion strings (outputs) given prompt strings (inputs). Prompt-based methods enable few-shot learning in LLMs for classification tasks \citep{brown2020}. By reformulating classification as a language modeling problem, the LLM predicts a target label token given a prompt template, and these approaches achieve competitive results with minimal labeled data \citep{pet1,singh2025}. Despite their success, studies reveal that prompts are often not interpreted by models in the way humans intend \citep{khashabi2022}, and they can inadvertently influence model biases in classification decisions \citep{prabhumoye2021,utama2021,webson2022,goral2025,chaudhary2025}.

From a taxonomic perspective, prompt-based classification methods can be broadly categorized into two lines: prompt design, which focuses on crafting human-readable prompts that map classification tasks to cloze-style questions for a frozen language model \citep{petroni2019}, and prompt tuning, where continuous soft prompts are learned while the underlying model remains fixed, enabling more flexible adaptation to target class distributions \citep{lester2021,qin2021,sanh2022}. In both paradigms, the core objective remains accurate classification across all classes—a goal that can be undermined when models exhibit systematic accuracy disparities between frequently and rarely occurring label tokens.

\subsection{Measuring Class Accuracy Disparities}
\citet{cobias} propose the COBias metric for evaluating pairwise class accuracy disparities, when using LLMs to perform text classification. COBias also enables learning debiasing coefficients that, during inference, plug in to reweight the probability distribution over the prediction token for classification tasks \citep{cobias,furud,dcs}. More recently, \citet{li2026} explored the use of an adapted Gini coefficient as a metric in multi-agent systems. In contrast, we treat Gini not as an adapted system-level metric but as a direct measure of class accuracy disparity within prompt-based classification, where it naturally captures relative imbalance across classes, enabling both interpretability and direct optimization.

% While the term ‘[keyword]’ has recently appeared in [other domain] (e.g., Author et al., 2025, rejected at ConfXYZ), its usage there focuses on [their shallow use]. In contrast, we treat it not as a metric but as a generative concept that helps uncover [your deeper mechanism].”

% % 如果担心被说“不 rigorous”，就加一句：
% We do not claim [concept] is measurable per se; rather, it serves as a heuristic scaffold to guide inquiry—consistent with its roots in [引用启发式理论，如 Kahneman & Tversky 或建构主义教育理论]。

\section{The Gini Index for Relative Class Accuracy Disparity in Prompt-Based Classification}
% \section{Gini Index in Classification Tasks}
\label{sec:gini}
We transfer Gini index, from the commonly used socioeconomic metric of income inequality, to a useful metric for class accuracy disparity in prompt-based classification. Below, we define the Gini index metric, illustrate why Gini makes sense with a numerical walkthrough, and compare Gini with a most related metric, COBias.
% 标。基尼系数最大为“1”，最小等于“0”。基尼系数越接近 0 表明收入分配越是趋向
% 平等。

%3.1 Gini Coefficient 定义
\subsection{Gini Index Definition}
The measurement of class accuracy imbalance strongly resembles inequality between strong and weak classes. Therefore, we can define Gini index as a metric of accuracy disparity in prompt-based classification results.

\noindent \textbf{The Reciprocal Calculation Method:}
Because prompt-based classification outputs often do not conform to standard parametric distributions, we adopt the direct, reciprocal Gini index method to measure overall class accuracy disparity. This method computes the Gini index without distributional assumptions---making the Gini Index a well-suited metric for analyzing prompt-based classification outputs.

We introduce the Gini index definition as follows. As a reminder, for \textbf{prompt-based classification}, an LLM works as modeling the probability distribution of the completion/answer string (the classification output) given a prompt consisting of task instructions, the instance to be classified, and a question. We follow a common practice to predict the \textit{argmax} class using probabilities assigned to label tokens. In details, after obtaining the answer string's probability distribution over the whole vocabulary, we extract and normalize probabilities corresponding to label tokens in the vocabulary to form class probabilities, i.e., $\boldsymbol{p}_m=(p_{m1},\dots, p_{mN})$ for instance $x_m$ and $N$ classes. The prediction $\hat{y}_m$ is then $\textit{argmax}_{i \in \{1,\dots,N\}} \boldsymbol{p}_{mi}$. We can then compute the accuracy for each class using the ground‑truth instances, to evaluate class-specific performance.

Concretely, let $A_i$ denote the accuracy for class $i, i \in {1,\dots, N}$. Let $\overline{\gamma}^{\text{Acc}}$ represent the average class accuracy:
\begin{equation}
    \overline{\gamma}^{\text{Acc}}=\frac{1}{N} \displaystyle \sum_{i=1}^{N} A_i
\end{equation}
As an analogy, $A_i$ resembles a family's income, and the average class accuracy is similar to average income over $N$ classes (families). Then, we can similarly define the Gini index about class accuracies in classification, which we term as \textit{Classification tasks Gini} or $G^{\text{CLS}}$, analogous to income Gini.

Henceforth, the $G^{\text{CLS}}$ metric is mathematically defined as follows.
\begin{equation}
    G^{\text{CLS}} = \frac{ \sum_{i=1}^{N} \sum_{j=1}^{N} |x_i - x_j| }{ 2N^2 \overline{\gamma}^{\text{Acc}} }
\end{equation}

% Gini index in classification tasks 
\noindent \textbf{Interpretation of the Gini Scale (Range: [0, 1]):}
\indent Gini = 0 indicates perfect fairness: all classes have equal accuracy; or in a perfectly equal society, the income difference is always 0. Gini approaches 1 (as population size approaches infinity) corresponds to maximal disparity: the accuracy gap between one class vs. the rest of all classes reaches the largest possible value; or in a perfectly unequal society (one person has everything), the average income difference approaches twice the mean\footnote{See Appendix \ref{appdix:derivation} for derivations}. Gini > 0.4 usually indicates strong relative imbalance \citep{jin2015}.

\subsection{A Numerical Walkthrough of the Gini Index Calculation}
Using the above definition, we present a step-by-step numerical demonstration of the Gini Index to illustrate how it works and why it makes sense. Below is the step-by-step breakdown of the $G^{\text{CLS}}$ formula.\\

\noindent \textbf{The Numerator:} $\sum_{i=1}^{N} \sum_{j=1}^{N} |x_i - x_j|$. This is the total sum of absolute class accuracy differences. For each class $i$, it calculates the absolute difference between its accuracy and every other class $j$'s accuracy. This sum captures total inequality in the population (classes). If every class had the same accuracy, the sum would be zero. The more spread out the accuracies, the larger this sum becomes.

\indent \textbf{Example A with 4 classes with accuracies: [1, 0, 0, 0]:}  Sum\textsubscript{A} = 3 + 1 + 1 + 1 = 6

\indent \textbf{Example B with 4 classes with accuracies: [0.8, 0.2, 0, 0]:}  
Sum\textsubscript{B} = 2.2 + 1 + 1 + 1 = 5.2

\indent \textbf{Example C with 4 classes with accuracies: [1, 1, 0, 0]:}  Sum\textsubscript{C} = 2 + 2 + 2 + 2 = 8

\citet{cobias} uses a similar numerator (the sum of only unordered distinct pairs ($i<j$)), which is divided by combination size $\binom{N}{2}$. This \textbf{mean absolute difference over all unordered pairs} forms the COBias metric. The advantage of COBias is the directness in representing inequalities between pairs of classes, but it omits normalization by mean accuracy, as we will see next.\\

\noindent \textbf{The Denominator:} $2N^2 \overline{\gamma}^{\text{Acc}}$. This denominator provides the \textbf{scaling by population and mean}. 

$N^2$: There are $N \times N = N^2$ pairs in the numerator's double sum. Dividing by $N^2$ turns the total sum into an average pairwise difference, i.e., the \textbf{mean absolute difference over all ordered pairs}.

$\overline{\gamma}^{\text{Acc}}$: Dividing by this mean accuracy makes the measure \textbf{scale-invariant}; this is what COBias lacks. The resulted $\frac{\text{average pairwise difference}}{\overline{\gamma}^{\text{Acc}}}$ is ``relative mean difference''. If every class accuracy doubled, $\overline{\gamma}^{\text{Acc}}$ would double and the sum would double, so the ratio stays the same. This is crucial---inequality shouldn't change just because everyone gets richer by the same proportion.

$2$: By conventional definition, the Gini index is half the relative mean difference \citep{gini1912}. The division by 2 is embedded in the normalization to bound the index between 0 and 1.

Putting it together, Gini indices for the following numerical examples are:

\indent \textbf{Example A with 4 classes with accuracies: [1, 0, 0, 0]:}  
$G^{\text{CLS}}_\text{A} = \frac{\text{Sum\textsubscript{A}}}{2 \times 4^2 \times 0.25} = \frac{6}{8} = 0.75$

\indent \textbf{Example B with 4 classes with accuracies: [0.8, 0.2, 0, 0]:}  
$G^{\text{CLS}}_\text{B} = \frac{5.2}{2 \times 4^2 \times 0.25} = \frac{5.2}{8} = 0.65$

\indent \textbf{Example C with 4 classes with accuracies: [1, 1, 0, 0]:}  
$G^{\text{CLS}}_\text{C} = \frac{\text{Sum\textsubscript{C}}}{2 \times 4^2 \times 0.5} = \frac{8}{16} = 0.5$

Intuitively, Gini index emphasizes how much the top class dominates relative to the mean, not absolute values. Gini is \textit{not} higher for larger absolute gaps, but when the top class exceeds the mean more. In summary, Gini (0.75 for [1, 0, 0, 0], 0.5 for [1, 1, 0, 0], 0.65 for [0.8, 0.2, 0, 0]) penalizes \textbf{proportional dominance}, not absolute gaps, so a split like [1, 1, 0, 0] (where top classes share dominance) scores lower than a monopoly [1,0,0,0]. Even though [0.8, 0.2, 0, 0] has a larger absolute gap (0.6) between top classes than [1, 1, 0, 0] (0), its Gini index being higher than [1, 1, 0, 0]'s is not just because of the larger gap, but because the top class (0.8) remains heavily dominant relative to the mean (3.2x mean), whereas [1, 1, 0, 0]'s top class has less relative concentration (2x mean).

\subsection{Comparisons Between Gini Index And COBias}
% The key difference is, Gini penalizes extreme concentration, while COBias penalizes widespread disparity. We will illustrate it with the above numerical cases. 
Both metrics as tools for different diagnostic or optimization goals. The core rationale is an optimization/bias mitigation target priority---Gini penalizes relative concentration (strong concentration if high inequality acro relative to the mean); COBias penalizes absolute pairwise gaps (large raw differences between classes), and which matters depends on the bias pattern of interest. 

Recall COBias definition: for class accuracies $\mathbf{A} = (A_1, \dots, A_N)$ across $N$ classes, the \textbf{COBias} metric is defined as the mean absolute difference over all distinct unordered pairs:
\begin{equation}
    \operatorname{COBias} = \frac{2}{N(N-1)} \sum_{1 \leq i < j \leq N} |A_i - A_j|.
\end{equation}
The mathematical relationship between Gini and COBias reveals their core difference: Gini measures \textbf{relative} accuracy imbalance (how accuracy is concentrated across classes relative to the mean), while COBias measures \textbf{absolute} accuracy imbalance.
\begin{equation}
    G^{\text{CLS}}_\text{C} = \frac{N-1}{2N\overline{\gamma}^{\text{Acc}}} \operatorname{COBias}.
\end{equation}
With our previous numerical examples, Table \ref{tab:comparison} compares values using Gini and COBias metrics.
\begin{table}[h]
\centering
\caption{Gini Index vs. COBias for three illustrative distributions}
\label{tab:comparison}
\begin{tabular}{lccc}
\toprule
\textbf{Eval. Metric} & [1, 0, 0, 0] & [0.8, 0.2, 0, 0] & [1, 1, 0, 0] \\
\midrule
Mean Acc. ($\overline{\gamma}^{\text{Acc}}$) & \textcolor{blue}{0.25} & \textcolor{blue}{0.25} & \textcolor{red}{0.50} \\
% \midrule
Gini ($G^{\text{CLS}}$) & \textcolor{red}{0.75} & 0.65 & \textcolor{blue}{0.50} \\
% \midrule
COBias & 0.50 & \textcolor{blue}{0.43} & \textcolor{red}{0.67} \\
\bottomrule
\end{tabular}
\end{table}

Table \ref{tab:comparison_summ} summarizes key differences between Gini index and COBias. The core insight is that, Gini is minimized when class performance is proportionally balanced; COBias is minimized when absolute differences between all class pairs are small.

\begin{table}[h]
\centering
\caption{Key differences between Gini index and COBias}
\label{tab:comparison_summ}
\rowcolors{1}{white}{cyan!15} % Alternating rows: 1st data row = white, 2nd = light cyan
\begin{tabular}{p{0.3\linewidth}p{0.3\linewidth}p{0.3\linewidth}}
\toprule
\textbf{Property} & \textbf{Gini Index} & \textbf{COBias} \\
\midrule
Normalization & Relative: divided by $2\overline{\gamma}^{\text{Acc}}$ & Absolute: no mean normalization \\
Scale dependence & Scale-invariant (homogeneous of degree 0) & Scale-dependent (homogeneous of degree 1) \\
Peak value condition & Maximal when one dominant class holds all mass ($A_k = \overline{\gamma}^{\text{Acc}} N$, others $=0$) & Maximal when values split between extremes (e.g., half at $\max$, half at $\min$) \\
Sensitivity & Relative inequality (proportional gaps) & Absolute dispersion magnitude \\
Interpretation & ``Fraction of mean separating random pairs'' $\times \frac{1}{2}$ & ``Expected absolute gap between random distinct pairs'' \\
Optimization Outcome & Gini is minimized when class performance is proportionally balanced & COBias is minimized when absolute differences between all class pairs are small.\\
\bottomrule
\end{tabular}
\end{table}

\noindent \textbf{What It Means:} from these comparisons, Gini scales disparities by the mean, making it invariant to overall magnitude and most sensitive to concentration, that is, dominant classes (e.g., [1,0,0,0] and [0.5,0,0,0] yield the same Gini). COBias computes the raw mean absolute difference over all unordered pairs, so it responds directly to the size of performance gaps, regardless of whether imbalance is concentrated in one class.

In essence, both metrics have their own advantages: Gini is well-suited when bias manifests as over-reliance on a dominant class, while COBias is more informative when large absolute discrepancies across multiple classes matter, even if none dominates.

% Here, COBias is higher not because disparity is "more widespread" (still concentrated in one dominant value), but because the absolute gap between 0.8 and 0.2 adds dispersion without changing the mean. This shows COBias responds to absolute differences, not "widespreadness" per se.

% [1,1,0,0] has 4 large gaps (all 1s), so its mean is high (0.667).
% [1,0,0,0] has 3 large gaps (all 1s), so its mean is lower (0.5).
% [0.8,0.2,0,0] has smaller gaps on average (0.6, 0.8, 0.8, 0.2, 0.2), so its mean is lowest (0.433).

\section{Benchmarking Gini Index of Class Accuracies in Prompt-based Classification Tasks}

We showcase Gini index evaluations in text and image classification scenarios. The task and model are randomly selected from existing literature, not cherry-picked. Besides Gini scores, we also report COBias scores as a baseline comparison metric.

\subsection{Case Analysis 1: Text Classification}
% 以文献[3]的 llama2-13b 产生的数据集 AGNews 为例，该任务分 4 类，各类的准确性及 Gini Coefficient 相关的数据见表一。
Using the Llama-2-13b few-shot text classification results on the dataset AGNews\footnote{We base our evaluations on the reported AGNews test set class accuracy of \citet{cobias}. Prompting details (e.g., prompt format, number of shots, etc.) can be found in the paper.} \citep{cobias} as a case study, we present the Gini index, alongside accuracy for each class, mean accuracy, top class relative dominance to the mean, COBias, and other measurement results in Table \ref{tab:res_text}. This task comprises 4 classes.

\begin{table}[ht]
\centering
\caption{Measurements of Gini index and related metrics for text classification (AGNews; 4 classes)}
\begin{tabular}{@{}lccccc@{}}
\toprule
\textbf{Eval. Metric} & \textbf{World} & \textbf{Sports} & \textbf{Business} & \textbf{Tech} & \textbf{Interpretation} \\ \midrule
\rowcolor[HTML]{DAE8FC} 
Class Acc. (4 Classes) & [0.85, & 0.98, & 0.97, & 0.19] & - \\
Mean Acc. ($\overline{\gamma}^{\text{Acc}}$) & \multicolumn{4}{c}{0.75} & \begin{tabular}[c]{@{}c@{}}Average performance \\ across all 4 classes.\end{tabular} \\
\rowcolor[HTML]{DAE8FC} 
\begin{tabular}[c]{@{}l@{}}Relative Dominance\\ of the Top Class\\ to the Mean ($\frac{A_{\text{max}}}{\overline{\gamma}^{\text{Acc}}}$)\end{tabular} & \multicolumn{4}{c}{\cellcolor[HTML]{DAE8FC}1.31} & \begin{tabular}[c]{@{}c@{}}Top class (0.98) is 1.31× \\ the mean, showing\\ weak relative dominance\end{tabular} \\
COBias & \multicolumn{4}{c}{0.42} & \begin{tabular}[c]{@{}c@{}}High value because \\ large absolute gaps exist\\ (e.g., 0.79 between 0.98 and 0.19), \\ so absolute disparity is significant.\end{tabular} \\
\rowcolor[HTML]{DAE8FC} 
{\color[HTML]{3531FF} Gini ($G^{\text{CLS}}$)} & \multicolumn{4}{c}{\cellcolor[HTML]{DAE8FC}{\color[HTML]{3531FF} 0.21}} & {\color[HTML]{3531FF} \begin{tabular}[c]{@{}c@{}}Low value (close to 0) because \\ the top class (0.98) is only \\ 1.31× the mean (not 4× like in {[}1, 0, 0, 0{]}), \\ so the system is relatively balanced\\ given the average performance.\end{tabular}} \\ \bottomrule
\label{tab:res_text}
\end{tabular}
\end{table}

\noindent \textbf{Interpretation of Scores:} in this text classification case, classes World, Sports, and Businesses are relatively strong with accuracies ranging from 0.85 to 0.98, while the weakest class Tech obtains a much lower accuracy 0.19. Gini (0.21) is low because the top class Sports's dominance (1.31× mean) is weak relative to the average performance, not because the class is balanced. As a comparison, COBias (0.42) is high because absolute gaps between classes (e.g., 0.79 between classes Sports and Tech) are large regardless of concentration.

As another case study, we use a relatively large LLM Llama-2-70b and obtainsfew-shot text classification results on the dataset DDI (drug-drug interaction relation classification;\citep{ddi}). We present the Gini index, alongside accuracy for each class, mean accuracy, top class relative dominance to the mean, COBias, and other measurement results in Table \ref{tab:res_text_ddi}. This task comprises 5 classes.

\begin{table}[ht]
\centering
\caption{Measurements of Gini index and related metrics for text classification (DDI; 5 classes)}
\begin{tabular}{@{}lcccclc@{}}
\toprule
\textbf{Eval. Metric} & \textbf{Neg.} & \textbf{Eff.} & \textbf{Mech.} & \textbf{Adv.} & \multicolumn{1}{c}{\textbf{Int.}} & \textbf{Interpretation} \\ \midrule
\rowcolor[HTML]{DAE8FC} 
Class Acc. (4 Classes) & [0, & 0.87, & 0.03, & 0.04, & 0.20] & - \\
Mean Acc. ($\overline{\gamma}^{\text{Acc}}$) & \multicolumn{5}{c}{0.23} & \begin{tabular}[c]{@{}c@{}}Average performance \\ across all 5 classes.\end{tabular} \\
\rowcolor[HTML]{DAE8FC} 
\begin{tabular}[c]{@{}l@{}}Relative Dominance\\ of the Top Class\\ to the Mean ($\frac{A_{\text{max}}}{\overline{\gamma}^{\text{Acc}}}$)\end{tabular}  & \multicolumn{5}{c}{\cellcolor[HTML]{DAE8FC}3.8} & \begin{tabular}[c]{@{}c@{}}Top class (0.87) is 3.8× the mean, \\ showing strong relative dominance.\end{tabular} \\
COBias & \multicolumn{5}{c}{0.38} & \begin{tabular}[c]{@{}c@{}}Moderate value because \\ large absolute gaps exist, \\ so absolute disparity is significant.\end{tabular} \\
\rowcolor[HTML]{DAE8FC} 
{\color[HTML]{3531FF} Gini ($G^{\text{CLS}}$)} & \multicolumn{5}{c}{\cellcolor[HTML]{DAE8FC}{\color[HTML]{3531FF} 0.67}} & {\color[HTML]{3531FF} \begin{tabular}[c]{@{}c@{}}High value (\textgreater 0.4) because \\ the top class is 3.8× the mean, \\ so the system is highly imbalanced\\ given the average performance.\end{tabular}} \\ \bottomrule
\end{tabular}
\label{tab:res_text_ddi}
\end{table}

\noindent \textbf{Interpretation of Scores:} The Llama-2-70b based DDI classification results show strong relative accuracy imbalance, and Gini is very high (0.67). Intuitively, the DDI accuracy distribution is visibly similar to the [0.8, 0.2, 0, 0] distribution in Table \ref{tab:comparison}, suggesting that even with a more comprehensive LLM, there can be extreme class accuracy disparities---dominance by a few classes.

\subsection{Case Analysis 2: Image Classification}
\begin{figure}[!t]
  \centering
  \includegraphics[width=0.8\linewidth]{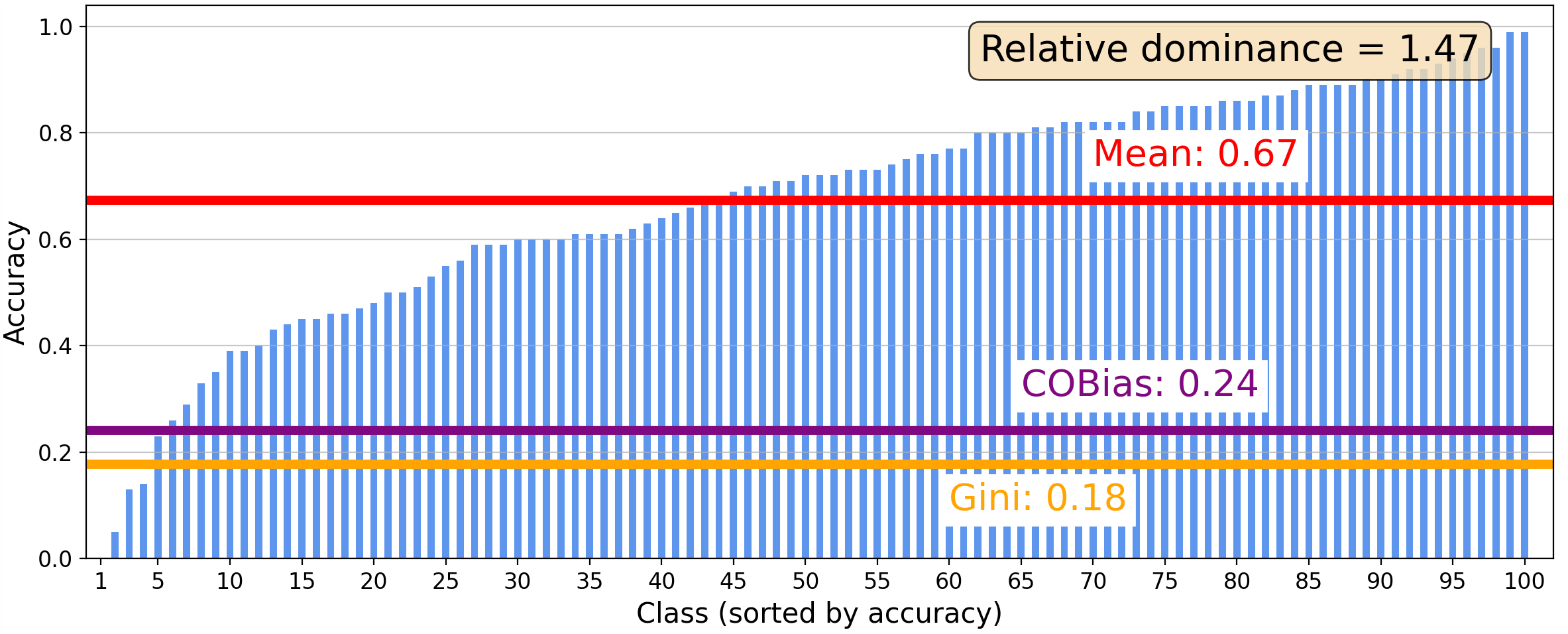}
  \caption{Measurements of Gini index and related metrics for image classification (CIFAR-100; 100 classes)}
  \label{fig:res_img}
\end{figure}

Using the CLIP ViT-L/14 zero-shot image classification results on the dataset CIFAR-100 \citep{cifar100} as a case study, we present the Gini index, alongside accuracy for each class, mean accuracy, top class relative dominance to the mean, COBias, and other measurement results in Figure \ref{fig:res_img}. This task comprises of 100 classes.

% [0.76, 0.92, 0.85, 0.82, 0.35, 0.85, 0.86, 0.61, 0.89, 0.75, 0.48, 0.29, 0.67, 0.77,  0.82, 0.93, 0.39, 0.5, , 0.61, 0.89, 0.99, 0.96, 0.88, 0.55, 0.87, 0.59, 0.82, 0.4,  0.45, 0.62, 0.69, 0.92, 0.53, 0.05, 0.86, 0.13, 0.82, 0.43, 0.8, , 0.6, , 0.68, 0.96,  0.66, 0.89, 0.7, , 0.26, 0.14, 0.85, 0.95, 0.73, 0.39, 0.73, 0.33, 0.84, 0.86, 0.5,  0.91, 0.59, 0.99, 0.7, , 0.23, 0.46, 0.9, , 0.51, 0.56, 0.71, 0.8, , 0., , , 0.85, 0.81,  0.72, 0.45, 0.6, , 0.6, , 0.61, 0.73, 0.61, 0.81, 0.84, 0.8, , 0.59, 0.77, 0.89, 0.94,  0.72, 0.87, 0.82, 0.8, , 0.65, 0.74, 0.9, , 0.63, 0.64, 0.71, 0.44, 0.76, 0.6, , 0.72,  0.46, 0.47]

\noindent \textbf{Interpretation of Scores:} for 100 classes with mean accuracy 0.67 (lowest class accuracy: 0, highest accuracy: 0.99), the low Gini index (0.18) confirms the absence of strong relative dominance---the top class is only 1.47× the mean accuracy. This demonstrates that Gini penalizes relative concentration (i.e., top class dominance relative to the mean), so a low score indicates no significant dominant class. COBias (0.24), meanwhile, reflects absolute gaps (e.g., 0.99 vs. 0), showing absolute accuracy disparity despite low relative concentration.

% Gini is not 0, confirming weak relative concentration exists. but it is insufficient to qualify as "dominant" (which would require Gini > 0.5, e.g., [1,0,0,0] → Gini=0.75)

\subsection{The Rationale}
In summary, analyses of these three randomly picked cases confirm that \textbf{weak to strong relative concentration exists in both prompt-based, text and image classification's class accuracies and regardless of whether the classification is high-dimensional or low-dimensional}. Crucially, this non-zero Gini (Gini > 0) demonstrates that \textbf{\textit{relative accuracy imbalance} is present in real-world models}. The DDI case exhibits a typical high Gini (0.67), calling for bias mitigation. Though non-dominant in AGNews and CIFAR-100 case studies (Gini = 0.18-0.21, where the top class is only 1.31-1.47× the mean accuracy), the weak concentration still represents measurable bias under our metric---such accuracy disparity must be mitigated.

\section{The Bias Mitigation Method}
Not only useful as an evaluation metric, Gini can be harnessed as an optimization metric for minimizing class accuracy disparities. We propose a bias mitigation method $D_{\text{Gini}}$ that directly leverages Gini. This method is post-hoc, model-agnostic, and without parameter updates of the original model.

The mitigation method presented here is a foundational proof of concept that directly validates the feasibility of mitigating relative accuracy imbalance. Crucially, the current method serves as a direct, unembellished translation of the proposed Gini metric into a functional bias mitigation model. While subsequent refinements will further optimize performance, this single model with experiments delivers clear, actionable proof that Gini-based mitigation is both possible and meaningful. 

\noindent \textbf{The Mathematical Model:} Given a classification dataset of $N$ classes, we split it into a labeled optimization set (analogous to the training set in gradient-based learning methods) of $M$ instances and a test set. After prompting the LLM/pretrained vision model for a prediction on an instance $x_m$ (from the optimization set), we can obtain the output class probabilities at the answer token: $\boldsymbol{p}_m=(p_{m1},\dots, p_{mN})$. The immediate prediction is $\hat{y}_m=\textit{argmax}_{i \in \{1,\dots,N\}} \boldsymbol{p}_{mi}$; and the per-class accuracy is
\begin{equation}
    A_{i}=\frac{1}{|\mathcal{S}_{i}|}\sum\nolimits_{m \in \mathcal{S}_{i}} \mathbbm{1}\{\hat{y}_m=y_m\},
\end{equation}
where $\mathcal{S}_{i}$ is the set of indices for class $i$ instances, and $y_m$ is the ground-truth class for instance $x_m$.

From the benchmarking results, we have seen that these prompt-based predictions can be prone to at least, weak relative accuracy imbalance (Gini > 0); therefore, we aim to post-hoc adjust the output class probability distribution, so that the corrected probabilities lead to fairer predictions and thus fairer class accuracies relative to the mean, i.e., lower Gini.

To post-hoc adjust the output class probability distribution for reducing Gini, we propose reweighting coefficients---specifically, integer selection variables $\boldsymbol{\xi} = (\xi_1,\dots,\xi_N)$ for class $\{1,\dots,N\}$ and a bias mitigation method $D_{\text{Gini}}$ based on $\boldsymbol{\xi}$---such that Gini index of the corrected predictions is minimized.

% Following the computational details of weight correction in \citep{cobias}, these $\boldsymbol{\xi}$ adjustments select from discrete, class-level correction weights (e.g., a factor of 0.2)---the correction is tailored to each class. We refer to the correction weights as the \textbf{correction map}: $\boldsymbol{F}=\{f_1,\dots,f_{|\boldsymbol{F}|}\}$. Henceforth, we correct the per-sample per-class probability $p'_{mi}$ by a mapping:

Following the computational details in \citep{dcs}, these $\boldsymbol{\xi}$ adjustments select from both discrete, class-level correction weights (e.g., a factor of 0.2) and sample-level correction functions (e.g., a probability-range-specific triangular membership function)---the correction is tailored to each class and instance. We collectively refer to the correction weights and correction functions as the \textbf{correction map}: $\boldsymbol{F}=\{f_1,\dots,f_{|\boldsymbol{F}|}\}$. Henceforth, we correct the per-sample per-class probability $p'_{mi}$ by a mapping:
\begin{equation}
    f_{\xi_i}: p'_{mi} \leftarrow f_{\xi_i}(p_{mi}), i \in \{1,\dots,N\}, p_{mi} \in [0,1], \xi_i \in \{1,\dots,|\boldsymbol{F}|\}.
\end{equation}
Therefore, the corrected prediction for instance $x_m$ is:
\begin{align}
\hat{y}'_m &= \textit{argmax}_{i \in \{1,\dots,N\}} \{p'_{m1},\dots,p'_{mN}\} \nonumber \\
&= \textit{argmax}_{i \in \{1,\dots,N\}} \{f_{\xi_1}(p_{m1}),\dots,f_{\xi_N}(p_{mN})\}.
\end{align}
Then, the updated accuracy for any class $i$ is
\begin{equation}
    A'_{i}(\boldsymbol{\xi})=\frac{1}{|\mathcal{S}_{i}|}\sum\nolimits_{m \in \mathcal{S}_{i}} \mathbbm{1}\{\hat{y}'_m=y_m\}.
\end{equation}
Within this notational framework, an optimization model (called $D_{\text{Gini}}$) aimed at minimizing Gini index in a prompt-based classification task can be formulated as follows:
\begin{align}
    \min G^{\text{CLS}}(\boldsymbol{\xi}) &= \frac{ \sum_{i=1}^{N} \sum_{j=1}^{N} |A'_{i}(\boldsymbol{\xi}) - A'_{j}(\boldsymbol{\xi})| }{ 2N^2 \overline{\gamma}^{\text{Acc}} } \nonumber \\ 
    \text{s.t.   } & \boldsymbol{\xi}=(\xi_1,\dots,\xi_N), \xi_i \in \{1,\dots,|\boldsymbol{F}|\}
\end{align}

\noindent \textbf{The Solution Framework Based on Simulated Annealing for the Mitigation Method}: to complete our solution framework for the mitigation model, without loss of generality, we follow \citep{dcs} to use a simulated annealing (SA) algorithm for solving the nonlinear integer programming objectives.

% \section{Discussion}

\section{Optimization Experiments}
% 优化实验
\subsection{Experimental Setup}
We apply $D_{\text{Gini}}$ to both case studies, including AGNews text classification and CIFAR-100 image classification. For optimizations on the CIFAR-100 dataset, we refine the correction map $\boldsymbol{F}$ to include only weight corrections. The simplification allows us to demonstrate the optimization performance while enabling rapid iteration for this case study. 

During inference, a test instance's class probabilities are reweighted by the reweighting coefficients for each class learned during optimization. We report evaluation scores on the test set. Evaluation metrics follow from the benchmarking section, including mean accuracy, Gini, COBias, and other related metrics.

\subsection{Bias Mitigation Results For Case Analysis 1: Text Classification}

Table \ref{tab:debiased_agnews} presents the test results on AGNews. Our bias mitigation method $D_{\text{Gini}}$ significantly reduces Gini index by relatively 86\%. In particular, top class Sports's relative dominance is reduced. Meanwhile, by optimizing over Gini, test COBias is also greatly reduced (by relatively 86\%), the weakest class Tech's accuracy rises from 19\% to 85\%. These suggest that by Gini-based bias optimization is effective for reducing both relative and absolute accuracy imbalance.

\begin{table}[ht]
\centering
\caption{Test results using the Gini-based bias mitigation method $D_{\text{Gini}}$ on AGNews (In the first column, $\uparrow$ ($\downarrow$) means higher (lower) value is better.)}
\begin{tabular}{@{}lccccccccc@{}}
\toprule
 & \multicolumn{4}{c}{\textbf{Original}} & \multicolumn{4}{c}{{\color[HTML]{3531FF} \textbf{Debiased (Opt. Metric: Gini)}}} &  \\
\multirow{-2}{*}{\textbf{\begin{tabular}[c]{@{}l@{}}Evaluation \\ Metric\end{tabular}}} & \textbf{World} & \textbf{Sports} & \textbf{Business} & \textbf{Tech} & \textbf{World} & \textbf{Sports} & \textbf{Business} & \textbf{Tech} & \multirow{-2}{*}{\textbf{\begin{tabular}[c]{@{}c@{}}Relative\\ Improvement\end{tabular}}} \\ \midrule
\rowcolor[HTML]{DAE8FC} 
Class Acc. (4 Classes) & 0.85 & 0.98 & 0.97 & 0.19 & \cellcolor[HTML]{D9FFC9}0.85 & \cellcolor[HTML]{D9FFC9}0.98 & \cellcolor[HTML]{D9FFC9}0.85 & \cellcolor[HTML]{D9FFC9}0.85 & \cellcolor[HTML]{FFFFC7}- \\
Mean Acc. ($\uparrow$) & \multicolumn{4}{c}{0.75} & \multicolumn{4}{c}{0.88} & $\uparrow$ 17\% \\
\rowcolor[HTML]{DAE8FC} 
\begin{tabular}[c]{@{}l@{}}Relative Dominance\\ of the Top Class\\ to the Mean ($\downarrow$)\end{tabular} & \multicolumn{4}{c}{\cellcolor[HTML]{DAE8FC}1.31} & \multicolumn{4}{c}{\cellcolor[HTML]{D9FFC9}1.11} & \cellcolor[HTML]{FFFFC7}$\downarrow$ 15\% \\
COBias ($\downarrow$) & \multicolumn{4}{c}{0.42} & \multicolumn{4}{c}{0.06} & $\downarrow$ 86\% \\
\rowcolor[HTML]{DAE8FC} 
{\color[HTML]{333333} Gini ($\downarrow$)} & \multicolumn{4}{c}{\cellcolor[HTML]{DAE8FC}{\color[HTML]{333333} 0.21}} & \multicolumn{4}{c}{\cellcolor[HTML]{D9FFC9}{\color[HTML]{333333} 0.03}} & \cellcolor[HTML]{FFFFC7}{\color[HTML]{333333} $\downarrow$ 86\%} \\ \bottomrule
\end{tabular}
\label{tab:debiased_agnews}
\end{table}

Table \ref{tab:debiased_ddi} shows the test results on DDI. Extremely imbalanced in the original prompt-based results, the corrected predictions yields much more balanced class accuracies. 

The saying---the minority classes usually offer the predictions that are most important---rings particularly true in the DDI classification, where mispredictions in a dominant class can lead to life-risking consequences (when patients trust the LLM predictions for a wrong interaction type between a pair of drugs and misuse them). \textbf{Our Gini-based bias mitigation method is particularly well-suited for biomedical DDI-like scenarios, effectively calibrating prompt-based outputs for much lower top class relative dominance}.

\begin{table}[ht]
\centering
\caption{Test results using the Gini-based bias mitigation method $D_{\text{Gini}}$ on DDI (In the first column, $\uparrow$ ($\downarrow$) means higher (lower) value is better.)}
\begin{tabular}{@{}lcccclcccclc@{}}
\toprule
 & \multicolumn{4}{c}{\textbf{Original}} &  & \multicolumn{4}{c}{{\color[HTML]{3531FF} \textbf{Debiased (Opt. Metric: Gini)}}} &  &  \\
\multirow{-2}{*}{\textbf{\begin{tabular}[c]{@{}l@{}}Evaluation \\ Metric\end{tabular}}} & \cellcolor[HTML]{FFFFFF}\textbf{Neg.} & \cellcolor[HTML]{FFFFFF}\textbf{Eff.} & \cellcolor[HTML]{FFFFFF}\textbf{Mech.} & \cellcolor[HTML]{FFFFFF}\textbf{Adv.} & \multicolumn{1}{c}{\cellcolor[HTML]{FFFFFF}\textbf{Int.}} & \cellcolor[HTML]{FFFFFF}\textbf{Neg.} & \cellcolor[HTML]{FFFFFF}\textbf{Eff.} & \cellcolor[HTML]{FFFFFF}\textbf{Mech.} & \cellcolor[HTML]{FFFFFF}\textbf{Adv.} & \multicolumn{1}{c}{\cellcolor[HTML]{FFFFFF}\textbf{Int.}} & \multirow{-2}{*}{\textbf{\begin{tabular}[c]{@{}c@{}}Relative\\ Improvement\end{tabular}}} \\ \midrule
\rowcolor[HTML]{DAE8FC} 
Class Acc. (4 Classes) & 0 & 0.87 & 0.03 & 0.04 & 0.20 & \cellcolor[HTML]{D9FFC9}0.30 & \cellcolor[HTML]{D9FFC9}0.45 & \cellcolor[HTML]{D9FFC9}0.32 & \cellcolor[HTML]{D9FFC9}0.27 & \cellcolor[HTML]{D9FFC9}0.52 & \cellcolor[HTML]{FFFFC7}- \\
Mean Acc. ($\uparrow$) & \multicolumn{4}{c}{0.23} &  & \multicolumn{4}{c}{0.37} &  & $\uparrow$ 61\% \\
\rowcolor[HTML]{DAE8FC} 
\begin{tabular}[c]{@{}l@{}}Relative Dominance\\ of the Top Class\\ to the Mean ($\downarrow$)\end{tabular} & \multicolumn{4}{c}{\cellcolor[HTML]{DAE8FC}3.8} &  & \multicolumn{4}{c}{\cellcolor[HTML]{D9FFC9}1.4} & \cellcolor[HTML]{D9FFC9} & \cellcolor[HTML]{FFFFC7}$\downarrow$ 63\% \\
COBias ($\downarrow$) & \multicolumn{4}{c}{0.38} &  & \multicolumn{4}{c}{0.13} &  & $\downarrow$ 66\% \\
\rowcolor[HTML]{DAE8FC} 
{\color[HTML]{333333} Gini ($\downarrow$)} & \multicolumn{4}{c}{\cellcolor[HTML]{DAE8FC}{\color[HTML]{333333} 0.67}} &  & \multicolumn{4}{c}{\cellcolor[HTML]{D9FFC9}{\color[HTML]{333333} 0.14}} & \cellcolor[HTML]{D9FFC9} & \cellcolor[HTML]{FFFFC7}{\color[HTML]{333333} $\downarrow$ 79\%} \\ \bottomrule
\end{tabular}
\label{tab:debiased_ddi}
\end{table}

\subsection{Bias Mitigation Results For Case Analysis 2: Image Classification}
Table \ref{tab:debiased_cifar} shows that our mitigation method effectively addresses relative class accuracy imbalances in zero-shot image classification, where the 100 classes obtain more balanced relative and absolute class accuracies. Specifically, Gini is reduced by a relative 61\%, and the weakest class accuracy rises from 0 to 27\%.

\begin{table}[ht]
\centering
\caption{Test results using the Gini-based bias mitigation method $D_{\text{Gini}}$ on CIFAR-100 (In the first column, $\uparrow$ ($\downarrow$) means higher (lower) value is better.)}
\begin{tabular}{@{}lclllclllc@{}}
\toprule
\textbf{\begin{tabular}[c]{@{}l@{}}Evaluation \\ Metric\end{tabular}} & \multicolumn{4}{c}{\textbf{Original}} & \multicolumn{4}{c}{{\color[HTML]{3531FF} \textbf{Debiased (Opt. Metric: Gini)}}} & \textbf{\begin{tabular}[c]{@{}c@{}}Relative\\ Improvement\end{tabular}} \\ \midrule
\rowcolor[HTML]{DAE8FC} 
Mean Acc. ($\uparrow$) & \multicolumn{4}{c}{\cellcolor[HTML]{DAE8FC}0.67} & \multicolumn{4}{c}{\cellcolor[HTML]{D9FFC9}0.69} & \cellcolor[HTML]{FFFFC7}$\uparrow$ 3\% \\
\begin{tabular}[c]{@{}l@{}}Relative Dominance\\ of the Top Class\\ to the Mean ($\downarrow$)\end{tabular} & \multicolumn{4}{c}{1.47} & \multicolumn{4}{c}{1.42} & $\downarrow$ 3\% \\
\rowcolor[HTML]{DAE8FC} 
COBias ($\downarrow$) & \multicolumn{4}{c}{\cellcolor[HTML]{DAE8FC}0.24} & \multicolumn{4}{c}{\cellcolor[HTML]{D9FFC9}0.10} & \cellcolor[HTML]{FFFFC7}$\downarrow$ 58\% \\
Gini ($\downarrow$) & \multicolumn{4}{c}{0.18} & \multicolumn{4}{c}{0.07} & $\downarrow$ 61\% \\ \bottomrule
\end{tabular}
\label{tab:debiased_cifar}
\end{table}

In summary, this mitigation method effectively harnesses Gini for meaningful bias mitigation. We leave the door open for iterative engineering or model enhancements given the fundamental validity of our approach.

\subsection{Discussion: Ablation On The Optimization Metric Choice, Gini. vs. COBias}
Through ablation, we isolate the impact of optimization metric choice: Gini vs. COBias. This COBias-based ablation study is performed on AGNews; hyperparameters align with the Gini-based setting. Results are shown in Table \ref{tab:ablation}, and it demonstrates that, Gini-based tuning yields quantitatively slightly stronger bias mitigation than COBias-based tuning.

\begin{table}[ht]
\centering
\caption{Ablation results using the COBias-based bias mitigation method $D_{\text{Gini}}$ on AGNews (In the first column, $\uparrow$ ($\downarrow$) means higher (lower) value is better.)}
\begin{tabular}{@{}lccccccccc@{}}
\toprule
 & \multicolumn{4}{c}{\textbf{Original}} & \multicolumn{4}{c}{{\color[HTML]{9A0000} \textbf{Debiased (Opt. Metric: COBias)}}} &  \\
\multirow{-2}{*}{\textbf{\begin{tabular}[c]{@{}l@{}}Evaluation \\ Metric\end{tabular}}} & \textbf{World} & \textbf{Sports} & \textbf{Business} & \textbf{Tech} & \textbf{World} & \textbf{Sports} & \textbf{Business} & \textbf{Tech} & \multirow{-2}{*}{\textbf{\begin{tabular}[c]{@{}c@{}}Relative\\ Improvement\end{tabular}}} \\ \midrule
\rowcolor[HTML]{DAE8FC} 
Class Acc. (4 Classes) & 0.85 & 0.98 & 0.97 & 0.19 & \cellcolor[HTML]{D9FFC9}0.86 & \cellcolor[HTML]{D9FFC9}0.98 & \cellcolor[HTML]{D9FFC9}0.85 & \cellcolor[HTML]{D9FFC9}0.84 & \cellcolor[HTML]{FFFFC7}- \\
Mean Acc. ($\uparrow$) & \multicolumn{4}{c}{0.75} & \multicolumn{4}{c}{0.88} & $\uparrow$ 17\% \\
\rowcolor[HTML]{DAE8FC} 
\begin{tabular}[c]{@{}l@{}}Relative Dominance\\ of the Top Class\\ to the Mean ($\downarrow$)\end{tabular} & \multicolumn{4}{c}{\cellcolor[HTML]{DAE8FC}1.31} & \multicolumn{4}{c}{\cellcolor[HTML]{D9FFC9}1.11} & \cellcolor[HTML]{FFFFC7}$\downarrow$ 15\% \\
COBias ($\downarrow$) & \multicolumn{4}{c}{0.42} & \multicolumn{4}{c}{0.07} & $\downarrow$ 83\% \\
\rowcolor[HTML]{DAE8FC} 
{\color[HTML]{333333} Gini ($\downarrow$)} & \multicolumn{4}{c}{\cellcolor[HTML]{DAE8FC}{\color[HTML]{333333} 0.21}} & \multicolumn{4}{c}{\cellcolor[HTML]{D9FFC9}{\color[HTML]{333333} 0.03}} & \cellcolor[HTML]{FFFFC7}{\color[HTML]{333333} $\downarrow$ 86\%} \\ \bottomrule
\end{tabular}
\label{tab:ablation}
\end{table}

\section{Conclusion}
We carefully examine Gini's hidden role as a tool for detecting and optimizing (debiasing) disparities in class accuracy, focusing on the case of prompt-based classification. In particular, we first empirically demonstrate Gini values in representative LLM based text and image classification scenarios---revealing that weak to strong relative accuracy imbalance exists in real-world models---and we directly leverage Gini to propose a post-hoc, model-agnostic bias mitigation method. Experimental results show that our method effectively reduces Gini from 0.21 to 0.03 for news text classification, from 0.67 to 0.14 for biomedical drug-drug interaction relation classification, and from 0.18 to 0.07 for a 100-class image classification task. This Gini-based mitigation method also significantly improves the weakest class's accuracy, showing that reducing relative accuracy imbalance is possible and meaningful.

It should be noted that, by proposing the Gini index and the post-hoc debiasing method in the prompt-based classification setting, we do not imply that they are only for LLMs, but rather because this is the high-impact context of accuracy disparity challenges. In addition, models examined in this work are basic yet widely-used architectures that serve as building blocks for more complex systems. Understanding and mitigating accuracy disparities in basic models is a necessary prerequisite before extending to multi-modal or multi-agent settings. Building on this foundation, we aim to extend our methods to multi-modalities and agentic systems in future work.

% \section*{Acknowledgments}
% This was was supported in part by......

%Bibliography
% \bibliographystyle{unsrtnat}  
\bibliography{output}  

\appendix

\section{Maximum Possible Gini Value $\frac{N-1}{N}$}
\label{appdix:derivation}
By the proposed definition, max Gini happens in the case of $[1, 0, \dots, 0]$ for $N$ classes. As $N \rightarrow \infty$, Average difference $ = \frac{2(N-1)}{N^2} \approx 2 \times \frac{1}{N}$, i.e, twice the mean. That is, when one class reaches the highest accuracy while others obtain 0, the average class accuracy difference approaches twice the mean.

\end{document}